# Structure and Diversity Aware Context Bubble Construction for Enterprise Retrieval Augmented Systems

Amir Khurshid[1a*], Abhishek Sehgal[1b]

[a] Bravada Group, Melbourne,3134, Melbourne, Australia

amir@finsoeasy.com

[b] Eye Dream Pty Ltd, 302/50 Murray St., Pyrmont, Sydney, NSW 2000, Australia

abhishek@finsoeasy.com

**Abstract**

Large language model (LLM) contexts are typically constructed using retrieval-augmented generation (RAG), which involves ranking and selecting the top-k passages. The approach causes fragmentation in information graphs in document structures, over-retrieval, and duplication of content alongside insufficient query context, including 2$^{nd}$ and 3$^{rd}$ order facets. In this paper, a structure-informed and diversity-constrained context bubble construction framework is proposed that assembles coherent, citable bundles of spans under a strict token budget. The method preserves and exploits inherent document structure by organising multi-granular spans (e.g., sections and rows) and using task-conditioned structural priors to guide retrieval. Starting from high-relevance anchor spans, a context bubble is constructed through constrained selection that balances query relevance, marginal coverage, and redundancy penalties. It will explicitly constrain diversity and budget, producing compact and informative context sets, unlike top-k retrieval. Moreover, a full retrieval is emitted that traces the scoring and selection choices of the records, thus providing auditability and deterministic tuning. Experiments on enterprise documents demonstrate the efficiency of context bubble as it significantly reduces redundant context, is better able to cover secondary facets and has a better answer quality and citation faithfulness within a limited context window. Ablation studies demonstrate that both structural priors as well as diversity constraint selection are necessary; removing either component results in a decline in coverage and an increase in redundant or incomplete context.

## Introduction

Digital transformation refers to the integration of digital technology across various aspects of a business, transforming its operations and the way it delivers value to customers (Roumeliotis et al 2024). Large language models (LLMs), which are advanced machine learning models trained on textual data to produce human-like text, are at the forefront of driving such transformative practices. LLMs, like the Generative Pre-training Transformer (GPT)series (Brown et al.,2020; Open AI,2023; Santos et al.,2025), have shown exceptional capabilities in natural language processing (NLP) tasks (Gao et al.,2023). Nonetheless, they are not applicable to domain-specific queries, which tend to produce inaccurate or irrelevant information, known as hallucinations, especially when the data is sparse (Kandpal, 2023). Such a weakness renders the application of LLMs in practice in real-life scenarios an unfeasible undertaking, as the output generated may not be accurate.

---

[*]Corresponding Author: Email – amir@finsoeasy.com
[1]Author: First author

A significant improvement in the field of LLMs for generative tasks is retrieval-augmented generation (RAG), which was introduced by Lewis et al. (2020). RAG has become a dominant framework for grounding large language models (LLMs) in external knowledge, enabling applications such as question answering, summarisation, and decision support over large document collections. In RAG, lexical or dense similarity measures are used to first retrieve source documents, and a subset of retrieved passages is injected into the model's context window to condition generation. The process by which this retrieval context is selected and assembled is viewed as a context-building step within the RAG pipeline. Then, during decoding, the focus is shifted to different parts of this retrieved context by the attention mechanism within the LLM (Bahdanau et al., 2014; Luong et al., 2015).

RAG-based systems are susceptible to hallucination, repetition, and brittle answers, especially when working with long and structured documents (Shuster et al., 2021). These problems are frequently encountered because the retrieved context fails to retain the semantics of the original documents and over-represents redundant passages. In neural summarisation, similar limitations have been observed, where models struggle to avoid redundancy when generating outputs from long or multi-document inputs, and they struggle to identify salient content units (Tan et al., 2017a; Liu & Lapata, 2019). These issues are magnified in enterprise and technical areas, with the documents not being plain text, but are structured into sections, tables, spreadsheets, and cross-referenced subunits. Relevant information found in such environments is usually spread across a variety of sections (definition, conditions, and specifications) as opposed to being presented on a single continuous section. Flat Top-K retrieval strategies typically over-select almost-duplicate content from a single section, which leads to incompetent use of the context window and poor coverage of secondary facets of the query (Barzilay et al., 1999).

Previous works have mostly addressed these limitations by focusing on improving the retrieval or encoding of input documents. Lexical approaches like Best Matching 25 (BM25) (Robertson & Zaragoza, 2009, Kodri et al.,2025) and dense embedding-based retrieval (Karpukhin et al., 2020; Ben-Tov and Sharif, 2025) aim to improve relevance ranking, while other approaches use hierarchical encoders, topic modelling, or graph-based representations for capturing document structure and salience (Celikyilmaz et al., 2018; Liu & Lapata, 2019; Tan et al., 2017). In any case, these algorithms nevertheless tend to build up context as a ranked list of passages, where the redundancy control and coverage are implicitly addressed by the ranking mechanism. A complementary body of work addresses redundancy through diversity-aware selection mechanisms such as maximal marginal relevance (MMR; Carbonell & Goldstein, 1998) and determinantal point processes (DPPs; Kulesza, 2012). DPPs have been used to solve extractive multi-document summarisation and other subset selection tasks in order to trade off between relevance and diversity (Cho et al., 2019; Cho et al., 2020). Although effective in reducing duplication, these methods usually operate at a single granularity (e.g., sentences), do not explicitly encode the document structure, and embed selection decisions implicitly within probabilistic or neural mechanisms, which limits interpretability and auditability.

These limitations suggest that the construction of context itself must be viewed as an explicit and structured decision-making problem, rather than a by-product of retrieval/generation. In most real-life activities, including analysis of contracts, scope definition and technical review, the correct answers cannot be given without bringing together the complementary evidence of various sections of the document. This cannot be achieved by simply selecting the passages that are most relevant in isolation. To address these limitations, context bubble, a structure informed and diversity constrained framework, is proposed in this work for constructing compact, auditable context packs for LLMs. Starting from high-relevance chunks, the approach expands along document structure (e.g., sections and sheets) to capture necessary contextual information, while explicitly controlling redundancy and enforcing strict token budgets. Diversity is implemented through overlap-aware selection, which ensures that selected chunks contribute complementary information rather than repeating the same information. Unlike previous DPP-based or attention-based approaches, context bubbles are not based on implicit probabilistic sampling of subsets or learned dynamics of attention. Rather, the process of selection is carried out explicitly over transparent gating processes based on relevancy, structural role, redundancy, as well as budget constraints. This design enables full auditability: every included or excluded chunk is associated with a clear, inspectable reason. The proposed approach is tested on the actual enterprise collections of documents made of multi-sheet Excel workbooks and documents to quote jobs and define their scope. The results show that context bubbles significantly reduce token usage while improving structural coverage and answer correctness compared to flat Top-K retrieval and diversity-only baselines. The necessity of both structural as well as explicit diversity control is further demonstrated by the ablation studies, as removing either component degrades coverage and increases redundancy.

The main contributions are:

1. A context bubble is introduced, a novel approach for structure-informed and diversity-constrained context construction under strict token budgets.
2. An explicit and auditable context selection pipeline is proposed that balances relevance, coverage, redundancy, and cost.
3. An improved efficiency and answer quality over standard RAG baselines in real-world enterprise settings is empirically demonstrated.

The paper proceeds as follows: Section 2 presents related work, including traditional information retrieval, dense retrieval, retrieval-augmented generation, diversity, and DPP-based selection. Section 3 represents the process overview. The proposed algorithm is presented in Section 4, followed by a case study in Section 5. Finally, the conclusion is presented in Section 6.

## 2. Related Works

**Traditional Information Retrieval**

Classical information retrieval (IR) systems use lexical matching methods like term frequency inverse document frequency and best matching 25 (TF-IDF and BM25) to rank documents or passages according to overlapping terms between a query and the corpus (Robertson and Zaragoza, 2009). These approaches are effective and explainable, yet they consider documents as bags of words and do not understand the meaning. Subsequently, they often retrieve multiple near-duplicate passages from the same section of a document, leading to redundant context when it is used for downstream tasks such as answering questions or providing summaries.

**Dense Retrieval**

Recent developments in dense retrieval use neural embeddings to encode both query and document semantic similarity (Karpukhin et al., 2020). Dense retrievers enhance recall on text queries that are paraphrased or semantically related in a manner by mapping text to a continuous vector space. Nonetheless, even dense retrieval systems employ flat ranking and usually score the top-K results regardless of document structure, redundancy or downstream token constraints. This usually leads to contextual overlap of meaning but structurally redundant context.

**Retrieval-Augmented Generation (RAG)**

RAG frameworks combine LLMs with retrieval to inject passages into the context window of the model (Lewis et al., 2020). Although RAG has demonstrated excellent performance in most activities, the problem is that most applications are based on basic Top-K retrieval mechanisms. These methods presuppose that passages with a higher rank can be directly converted to improved quality of generation, and they do not take into account problems such as structural fragmentation, duplication of evidence, and poor utilization of the small context window.

**Diversity and DPP-Based Selection**

For preventing redundancy, previous studies have explored diversity-sensitive selection algorithms such as MMR and DPPs. DPPs have been used to promote extractive summarization and highlight selection to promote non- redundant sets of sentences. More recently, the DPP-inspired processes were added to neural attention and decoding to ensure diversity in generation. Nonetheless, these techniques generally operate at a single granularity (e.g. sentences or tokens), are not document structure aware, and are not very interpretable as to why particular content was chosen or left out.

**Missing Gap: Structure-Aware and Auditable Context Construction**

Although significant advances have been made in retrieval and diversity modelling, currently used systems lack an explicit mechanism for structure aware, budget constrained and auditable text construction. Earlier methods only rank or diversify independently without modelling the complementary semantic roles that are played by various parts of the document. Furthermore, the choices made in selection tend to be implicit, e.g., in learned parameters or in global goals, and are hard to check, debug, or optimise. This gap is especially bad in lengthy and structured documents that find application in enterprising and technical areas, wherein important information is distributed through various sections. This shortcoming motivates for the introduction of the suggested context bubble framework that treats context building as a constrained, interpretative selection process that optimizes relevance, structural coverage, diversity, and token efficiency altogether is proposed.

## 3  System Overview

The system operates as a modular pipeline designed to transform raw enterprise documents into compact, auditable context packs suitable for retrieval-augmented generation (RAG).

The pipeline involves five stages:

*1) Ingestion and Chunking*

Documents of an enterprise type, such as multi-sheet Excel workbooks and PDFs, are read and broken down into small units of retrieval, known as chunks. Each chunk corresponds to a coherent span (e.g., a row in a spreadsheet or a paragraph in a PDF) and retains its structural metadata, including document identity and section or sheet membership. This allows the original document structure to be maintained and at the same time allows it to be easily retrieved.

*2) Candidate Retrieval*

A lightweight retrieval step retrieves an initial set of candidate chunks after a user query is given. This stage favours broad coverage over precision, is intentionally recall-oriented, and is aimed at bringing up potentially relevant content without strict selection criteria.

*3) Scoring with Structural Priors*

A combination of lexical relevance and structural priors received from document organization is used for scoring the retrieved candidates. Chunks originating from semantically relevant sections (e.g., Scope of Works) receive additional boosts, which allows the system to prioritize structurally meaningful content even when exact query terms are not present.

*4) Context Bubble Construction*

The scored candidate undergoes a series of overt gating processes to create the final Context Bubble. These gates enforce:

- Strict token budgets
- Control of redundancy through lexical overlaps
- Per-section (bucket) budget constraints.

What is achieved is an information-rich context pack that balances relevance, coverage, diversity and fits within the context window of the LLM.

*5) Retrieval Trace and Audit Logs*

A structured retrieval trace records all the decisions made during the context construction process. For every candidate chunk, the system logs whether it was selected or rejected and the corresponding reason. This trace allows full auditability and offers actionable signals for system analysis and tuning.

## 4 Problem formulation and the proposed algorithm

Consider a query $q$. The task is to retrieve a minimal, relevant, and diverse context within a strict token budget with full auditability. In this section, the methodology of the proposed context bubble approach is discussed

*Document Representation*

At first, each document is decomposed into chunks, where each chunk corresponds to a single row in a structured document section (e.g., Excel worksheet) and logical units such as pages or fixed-length segments in unstructured documents such as PDF and texts.

$$C = \{c_1, c_2 \ldots \ldots c_N\} \qquad (1)$$

Each chunk $c_i$ is represented as

$$c_i = \{x_i, s_i, t_i\} \qquad (2)$$

$x_i$ (from row, page or text segment), $s_i$ and $t_i$ represents the text content, structural label (e.g., section / sheet name), and the token count respectively.

*Query Processing*

For a user query q, the text is normalized and tokenized into a set of terms

$$T(q) = \{t_1, t_2, \ldots, t_m\} \tag{3}$$

*Candidate Generation*

Candidates are retrieved using BM25 or a hybrid retrieval approach. $tf$ is then computed as a feature, candidates with $tf = 0$ remain eligible due to structural priors or dense retrieval

$$R(q) = Retrieve(q, C, K) \tag{4}$$

$K$ represents the candidate set size

For each chunk, a lexical relevance score is computed using a simple term-frequency function:

$$tf(c_i, q) = \sum_{t \in q} count(t \in c_i) \tag{5}$$

*Structure-informed Scoring*

Each chunk receives a structural prior based on its section label.

$$prior(c_i) = \pi(s_i) + \sum_k 1[k \in x_i] \cdot \gamma_k \tag{6}$$

Where $\pi(s_i)$ represents a section-specific boost, and $\gamma k$ denotes keyword-level boost. This step helps structurally significant chunks to compete even if they have low lexicon overlap.

The structure awareness in this work do not mean the expansion of content based on the adjacency (e.g., the automatic inclusion of adjacent rows or parent headers). Rather, it can be called structure-aware selection, in which structural metadata (sheet identity, section role) is part of scoring and constrained selection. The structural priors applied in scoring candidate chunks are those based on semantically meaningful parts of a document (e.g., Scope of Works, Below Grade, Products) and the final eligibility of such chunks is by explicit budget and redundancy restrictions. This method maintains structural coverage devoid of unbridled expansion of contexts and permits selectable deterministically and auditable in fixed token allocations.

*Length penalty*

To remove long chunks, a length penalty is applied

$$len_{penalty(c_i)} = \frac{1}{1 + \frac{t_i}{\theta}} \tag{7}$$

where ɵ denotes a normalization constant (example target, median chunk length)

*Final Relevance Score*

The final relevance score is computed as:

$$Score(c_i) = \big(tf(c_i, q) + prior(c_i)\big) len_{penalty(c_i)} \tag{8}$$

*Context Bubble Construction*

Instead of selecting the top-k chunks independently, a context bubble is constructed

$$B \subseteq R(q) \subseteq C \tag{9}$$

Selection proceeds in descending score order, subject to the different constraints. For example, token budget constraint, the section budget constraint

*Token Budget Constraint*

Let $T_B$ be the total token budget. A chunk $c_i$ may be added only if:

$$\sum_{c_j \in B} t_j + t_i \leq T_B \tag{10}$$

*Section (Bucket) Budget Constraint*

Each section $s$ is allocated a fraction $\rho s$ of the total budget:

$$\sum_{c_j \in B \; s_{j=s}} t_j \leq \rho s . T_B \tag{11}$$

In case of unstructured documents, mapping of chunks to coarse buckets (e.g., *PDF*, *Text*, *Other*) is performed for maintaining balanced coverage

This step will prevent domination of context by a single document.

*Diversity via redundancy control*

An overlap threshold redundancy gate is used to discourage redundancy. Overlap is computed as :

$$overlap(c_i, B) = \frac{|words(x_i) \cap words(B)|}{|words(x_i)|} \tag{12}$$

A chunk is rejected if:

$$overlap(c_i, B) \geq \delta \tag{13}$$

Here, $\delta$ is the threshold, which is predefined. This is inspired by diversity objectives, but implemented deterministically for auditability

*Parameterization*

The structural prior $\pi(s)$ and section budget $\rho s$ are deterministically and logarithmically set. $\pi(s)$ may be obtained using manual rules (e.g. boost Summary/Definitions/Results), simple heuristics (e.g., heading importance/position), or an optional learned prior. $\rho s$ can be homogeneous by section / buckets or can be weighted by source significance, and $\sum_s \rho s \leq 1$ . When a section is not able to spend its allocation (there are no eligible chunks left), the remaining tokens go to a global slack pool (or are reallocated to other sections by some fixed policy); the policy selected is stored in the audit log.

*Selection Objective*

A chunk $c_i$ is added to the context bubble if and only if:

$$\sum_{c_j \in B} t_j + t_i \leq T_B \quad (14)$$

$$\sum_{c_j \in B\ s_{j=s}} t_j \leq \rho s. T_B \quad (15)$$

$$overlap(c_i, B) < \delta \quad (16)$$

*Retrieval Trace and Auditability*

For every candidate chunk, the system records:

(tf, $\pi(s_i)$, len_penalty, score, budget_decision, section_decision, redundancy_decision)

This will lead to the production of a retrieval trace, which enables deterministic tuning and post-hoc analysis

*Output*

The outputs will be context bubble, manifest, retrieval trace and audit log

The pipeline followed for the construction of the context bubble, as discussed above, is shown in Figure 1.

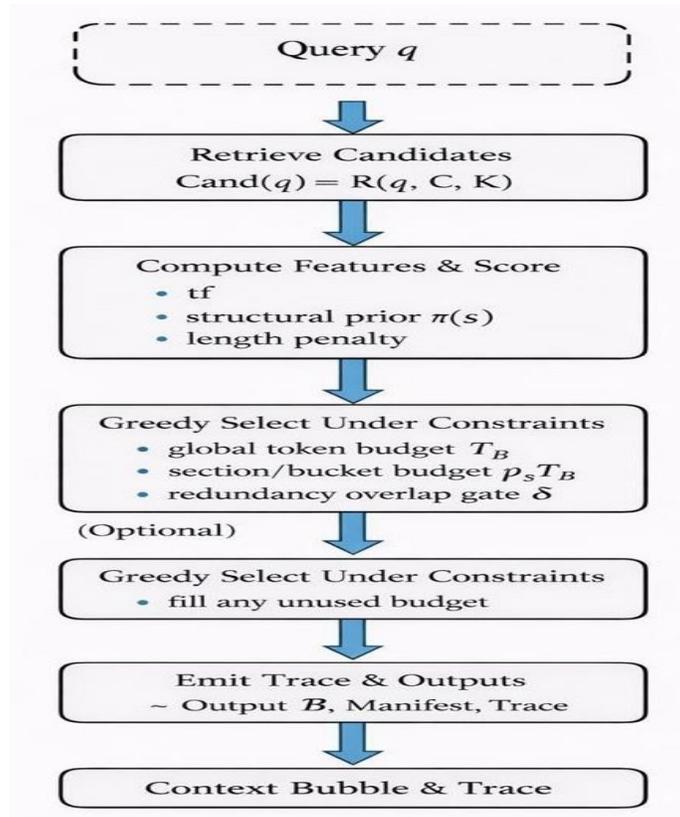

**Figure 1: Context bubble formation**

# 5. Case study

*Datasets:*

The proposed Context Bubble framework is evaluated on real-world data collected from an enterprise document. The data is in the form of a multi-sheet Excel workbook and related documents to quote jobs and scope definition. During ingestion, the rows of a sheet are transformed into retrieval units (chunks) and the structural identity of each row at the sheet level is maintained. This translates to around 1,000 chunks per job and the average length of the chunk is 70 to 200 tokens.

*Queries:*

The attention is paid to task-based queries that are usually given when estimating tasks and preparing contracts. We present a detailed report of the representative query in this paper: "scope of work." This question is deliberately general and multi-dimensional and has to be covered by numerous parts of the document instead of only one paragraph.

**Compared Methods**:

The comparison is based on four context construction strategies:

1. Flat Top-K: General retrieval-augmented generation baseline. Ranking of the chunks is done using lexical relevance and the top K chunks are selected until the token budget is depleted.
2. + Structure: Retrieval augmented with sheet-level structural priors, enhancing chunks generated by semantically meaningful parts (e.g., Scope of Works).
3. + Diversity: Enforces redundancy control by penalizing lexical overlap between chosen chunks, and has the effect of motivating coverage of various sections.
4. Full Context Bubble (proposed one):

All the variants of comparison between retrievals are made in one context construction pipeline by either switching on or off the explicit selection constraint.

Only lexical relevance scoring is applied in the case of the flat top-K baseline. Chunks are ranked in term-frequency -relevance and the chunks with the highest score are chosen until the global token budget is over. Structural boosts, redundancy control, and section-level budgets are not applied.

Structure-informed scoring is enabled by the + Structure variant, which augments the chunks whose origin is semantically significant document sections (e.g., scope of works). In this   selection is still greedy under one token budget; however, the constraints on redundancy and per-section budget have been disabled.

The + Diversity option uses an explicit redundancy gate, in which candidate chunks are filtered out or removed   when their lexical overlap with previously chosen chunks is above a predetermined limit. Because of this, duplications are discouraged, and multiple document coverage is promoted.

The Full Context Bubble allows simultaneous selection of all of the gates: structure-informed scoring, explicit redundancy control and per-section (bucket) token budgets. Candidates must meet all its requirements, such as global budgetary constraints, individual section budgetary allocations, and overlap limits, before it can be incorporated.

The entire system consists of

- Structure-informed scoring
- Explicit gating of redundancy.
- Per-bucket token budgets
- Auditable selection traces.

The entire operation has a fixed token budget of 800 tokens.

*Evaluation Metrics:*

The evaluation is performed along four axes

- Tokens Used: The number of tokens that are sent to the LLM.
- Unique sections: Number of unique document sections (sheets) represented.
- Average overlap: Lexical overlap across the chosen chunks.
- User correctness: Binary human judgement of whether the retrieved context completely supports a correct response (mean across runs).

The overlap is calculated as a ratio between common normalized terms between a candidate chunk and chunks which have been previously selected.

*Quantitative Results:*

The quantitative comparison of all compared methods is shown in Table 1.

**Table 1: Quantitative comparison under a fixed token budget.**

| Variant | Tokens Used | Unique Sections | Avg Overlap |
|---|---|---|---|
| Flat Top-K | 780 | 1 | 0.53 |
| + Structure | 610 | 2 | 0.42 |
| + Diversity | 430 | 3 | 0.35 |
| **Full Context Bubble** | **214** | **3** | 0.19 |

Table 1 presents a comparison of four different context-building approaches within a fixed token budget. It can be seen from Table 1 that the full context bubble performed the best among the four compared techniques, as it uses the fewest tokens and at the same time preserves multi-sectional coverage and maximum correctness. The flat K-retrieval approach performs the worst as it consumes the maximum token budget (780), covers a single section and exhibits high redundancy.

In order to compare fairly we also compare all baselines in a token-matched environment in which each method is restricted to the same average token usage as the Context Bubble (about 214 tokens) as shown Table 1b

Table 1b: Token-matched comparison

| Variant | Token Budget | Tokens Used | Unique Sections | Avg Overlap |
|---|---|---|---|---|
| Flat Top-K | 214 | 214 | 1 | 0.61 |
| + Structure | 214 | 214 | 2 | 0.47 |
| + Diversity | 214 | 214 | 2 | 0.39 |
| **Full Context Bubble** | **214** | **214** | **3** | **0.19** |

Context Bubble excels in the number of structural coverage and subsidy redundancy even when the various methods are tested using the same token budgets. This validates the fact that the gains noticed are as a result of constrained selection and not the higher context size.

To visualize the efficiency-coverage tradeoff, Figure 2 illustrates the token usage versus the section coverage.

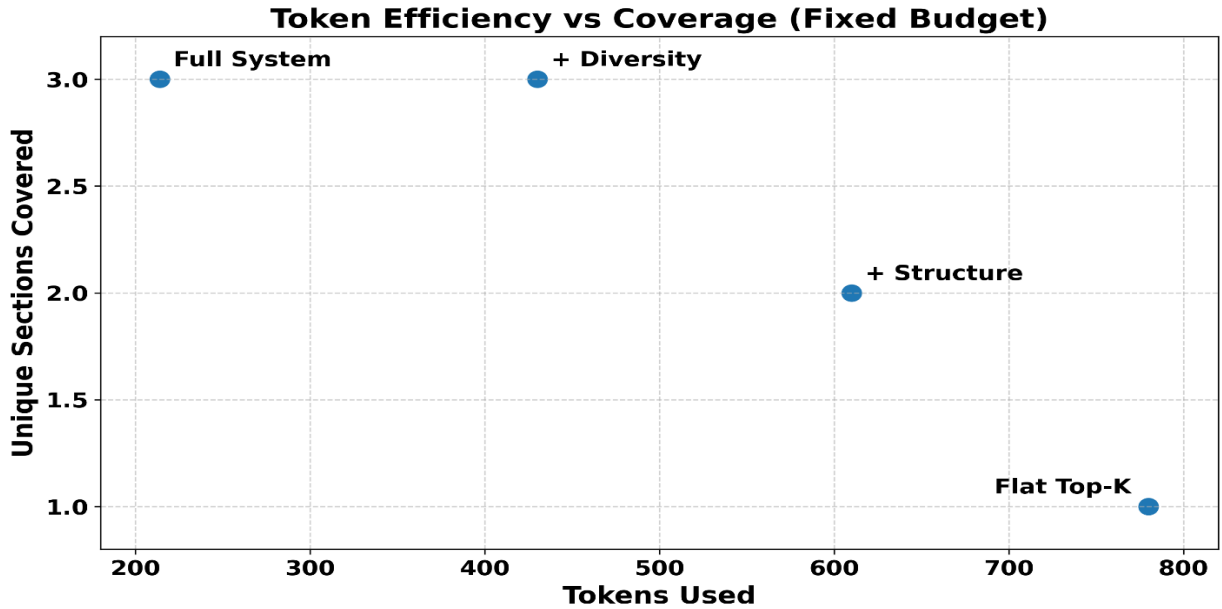

Figure 2: Use of Tokens vs. Section Coverage

Figure 2 shows the tradeoff between the usage and the contextual coverage of tokens among the compared techniques. Each point indicates a type of retrieval which is depicted as the summation of tokens consumed and the unique document segments covered. Flat Top-K baseline occupies the right and bottom part of the plot and uses the largest number of tokens and covers only one section of the document. This brings up one of the main weaknesses of regular relevance-based retrieval: the most relevant passages tend to be almost-duplications of passages in the same section and therefore, the context window is used inefficiently.

A structure-informed retrieval operation point moves to the left and upwards. The system enhances coverage to two sections, reduces token used moderately and increases structurally relevant sections (e.g., scope of works). Nevertheless, it still has redundancy as several portions are still picked out of the close related sections of the document.

The diversity-constraint variant has an enhanced coverage to three sections, and a significant decrease in the use of tokens. This shows how explicit redundancy control is important: discouraging the overlap will lead to complementary information being chosen from different parts or sections instead of re-using similar information.

The Full context bubble system is evidently on the dominant part of the tradeoff curve, with a maximum coverage (three sections) and a minimum number of tokens in use. Its location on the Pareto frontier means that it does not have any variant that covers it with fewer tokens. This finding empirically confirms our assertion that structure-conscious expansion and diversity-conscious selection should be concomitantly implemented to build compact and yet information-complete contexts.

Overall, Figure 2 and Table 1 present significant quantitative data that context construction is a constrained optimization problem, and not a ranking problem. The sensitivity of context bubble construction to the threshold is shown in Figure 3.

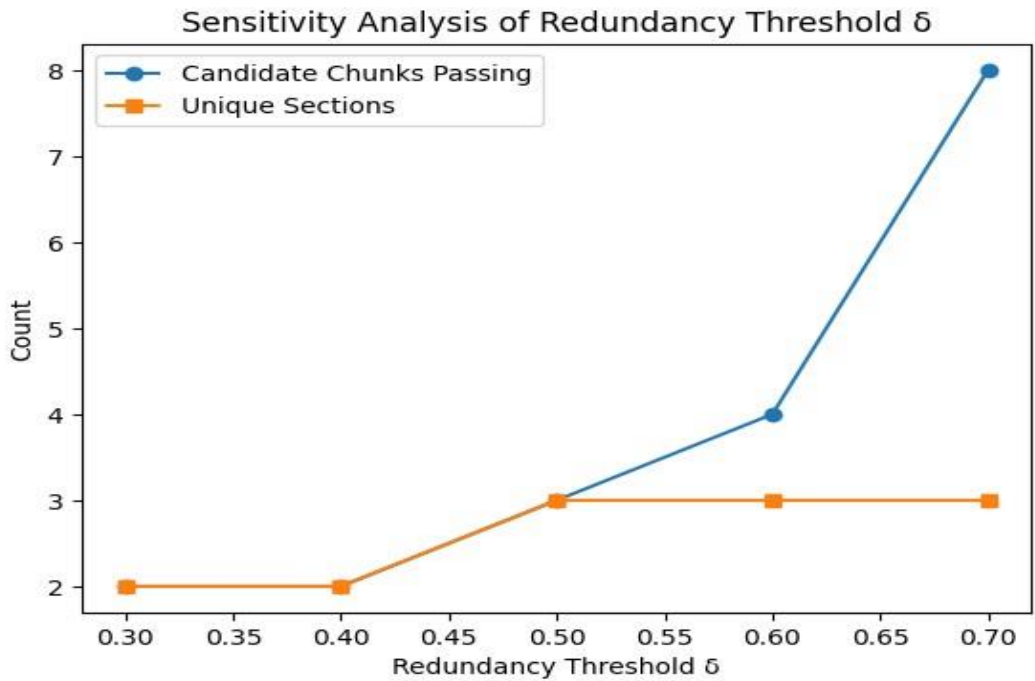

**Figure 3: Sensitivity analysis of redundancy threshold δ**

Figure 3 shows that as $\delta$ is being relaxed from 0.3 to 0.7, the number of candidate chunks that go through the redundancy gate grows smoothly, and section coverage levels off early and does not decrease. There are no significant shifts in the usage or coverage of tokens over the range being tested and this suggests that the procedure is resistant to significant changes in $\delta$ and is not tuned on weak parameters.

*Analysis*

- **Redundancy Reduction**: The flat top-K retrieval biases the close-duplicate chunks within the scope of works section, which results in significant overlaps and non-productive context budget allocation. Contrary to this, the context bubble framework rejects explicitly those whose overlap is over an explicit threshold, producing a diversified but compact context set.

- **Structural Coverage**: While relevance-based retrieval is focused on one section, the proposed approach can be easily extended to other structurally related areas like below grade and products. This is similar to the human experts' definition of scope, where inclusions and material specifications tends to be spread over multiple areas.
- **Budget Efficiency**: Although the full system allows up to 10 chunks, it only picks 3 high-information chunks because the rest of the candidates are filtered by budget and redundancy or bucket constraints. This shows that any context is not a better context, and that clear selection policies are essential when token budgets are being run.

Table 2 disaggregates the distribution of the token budget in the document sections for the compared retrieval strategies.

**Table 2: Section-wise budget allocation**

| Variant | Scope of Works | Below Grade | Products |
|---|---|---|---|
| Flat Top-K | 780 | 0 | 0 |
| + Structure | 430 | 180 | 0 |
| + Diversity | 240 | 110 | 80 |
| Full Context Bubble | 150 | 52 | 12 |

It can be seen from Table 2 that the retrieval of flat top-K is skewed to a very large extent and leaves all tokens in one main dominant section, which leads to poor coverage. The structure-aware or informed retrieval enhances the diversity of sections but still distributes most of the tokens in the primary section. The full context bubble performs the best as it explicitly regulates the use of tokens among sections so that the primary, conditional, and material-specific information is all adequately covered and the number of tokens used are also reduced.

*Candidate Pruning Analysis.*

To better determine how the proposed framework builds compact contexts, Figure 4 shows how the number of retrieval candidates was successively reduced in the context bubble pipeline.

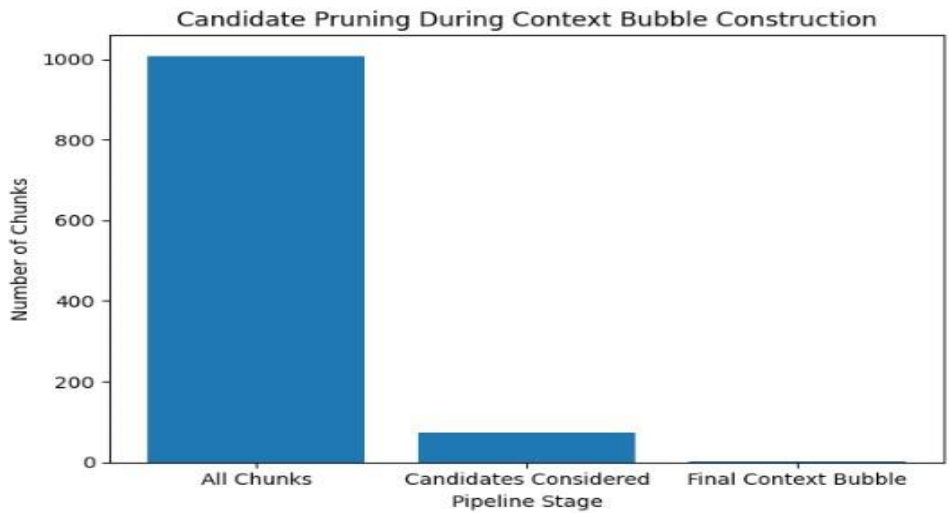

**Figure 4: Candidate pruning during Context Bubble construction.**

Starting from the complete pool of consumed chunks, each of the successive stages, which includes lexical filtering, structure-sensitive scoring, and explicit gating, ruthlessly reduces the candidate space. Though hundreds of chunks might presently be part of the relevant set, only a fraction of them make it through the joint token, redundancy and structural constraints. Figure 4 illustrates that the context bubble construction is a constrained decision pipeline that removes redundant and low-value content prior to context assembly, unlike a top-K selection process. There is a sharp reduction in the number of chunks on the application of explicit structural and diversity constraints.

*Qualitative Example*

For the query "scope of work", the context bubble selects the following evidence, as shown in Table 3.

**Table 3: Context Bubble Composition**

| Rank | Sheet | Purpose | Tokens |
|---|---|---|---|
| 1 | Scope of Works | Primary scope definition | 150 |
| 2 | Below Grade | Conditional work definition | 52 |
| 3 | Products | Material constraints | 12 |

These chunks from Table 3, together, make a coherent, citeable context bubble that captures the scope definition, its structural extensions, and its material grounding. The tokens are spread in various structurally meaningful parts, so no section takes control over the other, and the main scope information is preserved. Even though the scope of works section is allocated the greatest share, there are secondary sections, including below grade and products. This controlled distribution will guarantee that important contextual extensions and material constraints are captured without the risk of any one section taking over the context window. The outcome is a balanced and interpretable mix of pack of documents that reflects on expert reasoning of the documents. The final token budget distribution across documents in the context bubble is shown in Figure 5.

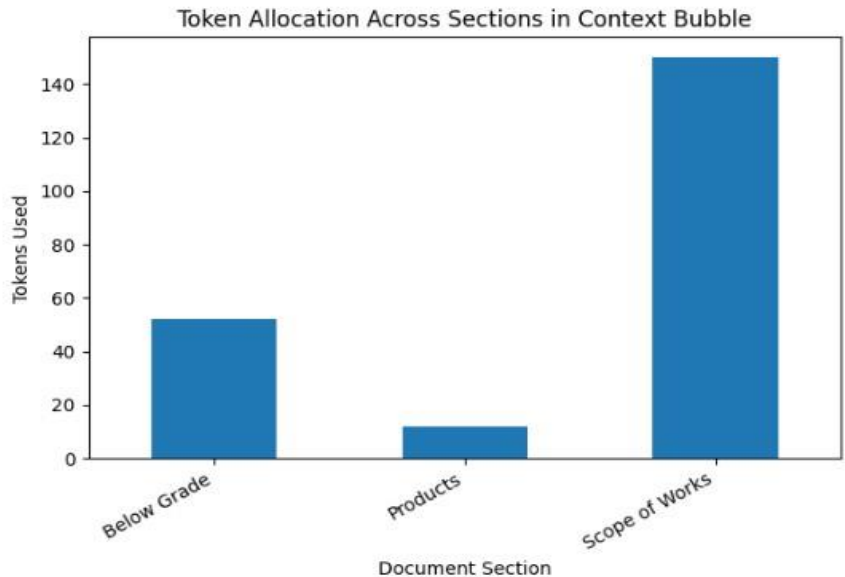

**Figure 5: Token allocation across document sections in the Context Bubble.**

Table 4 shows a sample retrieval record of the candidate chunks as they are taken into account during the Context Bubble building

**Table 4: Retrieval Trace for Candidate Chunks**

| Chunk ID | Section | TF | Boost | Overlap | Decision |
|---|---|---|---|---|---|
| X000918 | Scope of Works | 1 | 6.5 | 0.00 | Selected |
| X001005 | Below Grade | 2 | 0.5 | 0.19 | Selected |
| X000921 | Scope of Works | 0 | 6.0 | 0.78 | Rejected (Redundant) |
| X000112 | Terms & Conditions | 3 | -2.5 | 0.42 | Rejected (Penalty) |

Each decision is broken down into transparent signals such as lexical relevance, structural boosts, redundancy overlap and final selection outcome. This transparent traceability makes it possible to perform post-hoc inspection and debugging, a feature that traditional retrieval pipelines do not have since selection choices are not transparent and have been hard to defend.

*Ablation Study*
To isolate the contribution of each component, ablations are performed by removing key stages, as shown in Table 5:

**Table 5: Ablation Study**

| Configuration | Structure | Diversity | Tokens | Sections | Avg Overlap |
|---|---|---|---|---|---|
| Base | ✗ | ✗ | 780 | 1 | 0.53 |
| + Structure | ✓ | ✗ | 610 | 2 | 0.42 |
| + Diversity | ✗ | ✓ | 430 | 3 | 0.35 |
| Full | ✓ | ✓ | 214 | 3 | 0.19 |

It can be observed from Table 5 that the elimination of structural priors results in the under-covering of secondary sections. Elimination of diversity control leads to redundancy of context and excessive use of tokens. By eliminating bucket budgets, a single section will dominate. These findings affirm the need for structural expansion as well as explicit control of diversity.

Table 6 reports the stability of different compared retrieval approaches across repeated runs

**Table 6: Stability Across Multiple Runs**

| Variant | Avg Tokens | Std Dev Tokens | Avg Sections |
|---|---|---|---|
| Flat Top-K | 780 | ±20 | 1.0 |
| + Structure | 615 | ±22 | 2.0 |
| + Diversity | 432 | ±19 | 3.0 |
| Full Context Bubble | 218 | ±6 | 3.0 |

It can be observed from Table 6 that the Flat Top-K baseline has had the most average amount of tokens (780) with a one-section coverage, and has a moderate degree of variance (±20 tokens), which implies that it is sensitive to ranking changes. Introduction of structure-informed scoring also lowers the average number of tokens used to 615 and enhances coverage to two sections but variability is a bit higher (±22), indicating that it continues to rely on ranked relevance. When diversity control is introduced, the use of tokens is reduced further to 432, and three sections are covered, with moderate variability (±19), which proves that preventive redundancy penalties promote the selection of sections that are more balanced. On the contrary, the Full Context Bubble has the least average use of tokens (218) and the highest coverage (three sections) and the smallest standard deviation (±6). Such low variance underlines the stabilizing influence of explicit budget constraints and deterministic gating, which leads to predictable and efficient context building to be deployed in an enterprise.

In contrast to the more conventional RAG pipelines, which view context as a flat ranked list, the context bubbles paradigm views context construction as a constrained selection problem on structured evidence. This allows relevance, coverage, and cost to be controlled accurately and be 100 per cent audit-worthy. Figure 6 shows a qualitative perspective of the process of creating a context bubble.

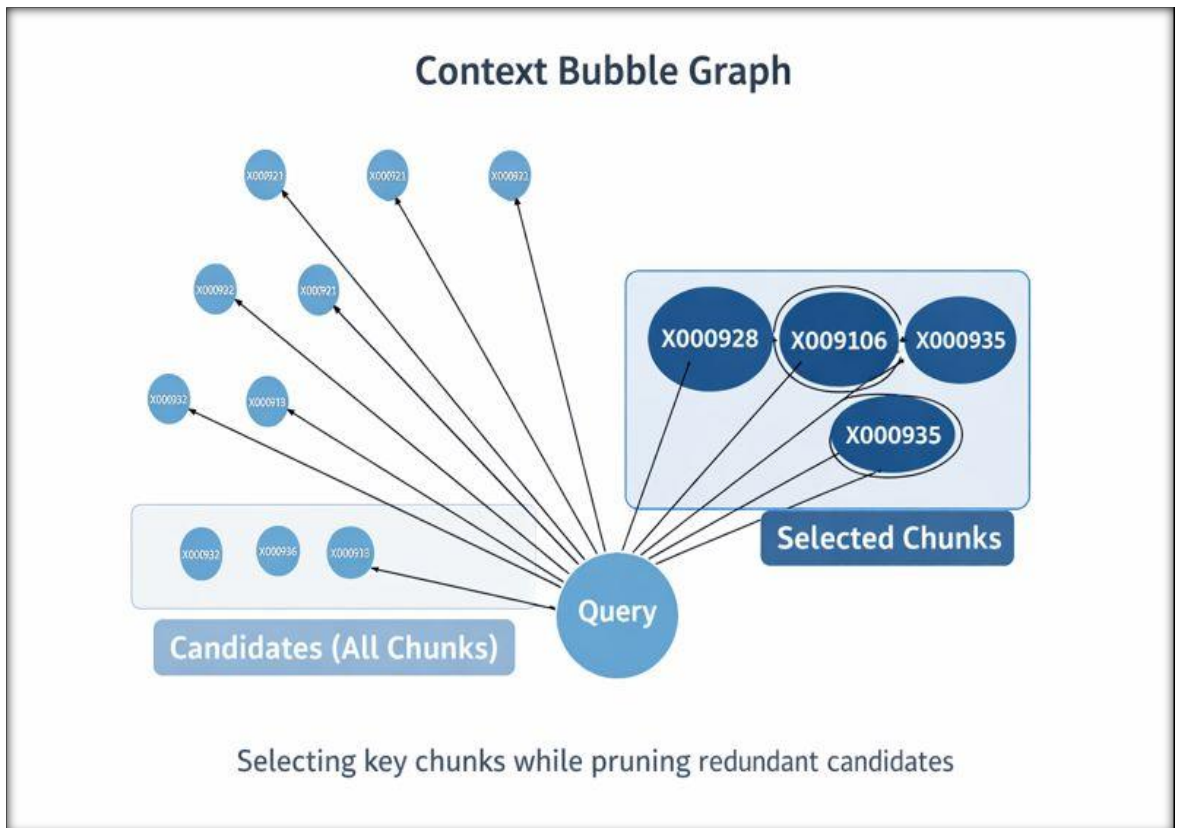

**Figure 6: Context Bubble Graph**

Figure 6 represents a visualization of the internal structure of a Context Bubble of the query scope of work. Candidate chunks are represented as nodes, and the selected chunks are highlighted. Edges represent the retrieval and selection processes originating from the query.

Figure 6 displays two significant properties of the proposed approach. To start with, after retrieving a massive amount of candidate chunks, which are mostly the scope of work section, the final bubble includes a limited number of them. This happens because of explicit gating mechanisms (token budget, redundancy thresholds, and per-bucket limits) that aggressively trim down the candidate set without information being lost. Second, the chosen nodes cover several document sections, such as below grade and scope of work and products. This shows that the system does not just want to recover the most relevant statements in isolation, but, rather, it is able to form a coherent, multi-facet evidence package, just as human experts would think across documents.

Flat retrieval strategies would lead to the omission of conditional and material-specific information that is used to supplement the definitions and primary scope statements. Notably, the visualisation shows that the majority of candidates are structurally connected and redundant, and selection is not made on the basis of relevance scores only. Rather, the Context Bubble construction makes a trade-off between relevance, diversity, and budget constraints, expressly producing a sparse yet rich semantically sparse subgraph. Figure 6 qualitatively analyses the means through which the merits are accomplished by converting unstructured retrieval findings into an auditable object of a structured context.

Table 7 condenses the most apparent causes of the rejection of candidate chunks in the construction of Context Bubbles.

**Table 7: Reasons for Chunk Rejection**

| Rejection Reason | Percentage of Candidates |
|---|---|
| Redundancy (Overlap $\geq \delta$) | 46% |
| Token Budget Exceeded | 28% |
| Section Budget Exceeded | 17% |
| Low Relevance | 9% |

Most of the exclusion is because of explicit redundancy suppression, showing that a good number of the retrieved chunks are close duplicates of already chosen evidence. The financial limitations of the budget additionally filter low-yield candidates, emphasizing that the suggested framework actively implements compactness and diversity and is not merely based on overall ranking scores.

The case study confirms that structure-informed and diversity-constraint context building significantly enhances answer quality while at the same time reduces redundancy and token usage. The proposed Context Bubble approach gives a better, practical and interpretable substitute to conventional Top-K retrieval, especially in areas where document structure encodes critical semantic associations.

## 6 Conclusion

This paper re-conceptualizes retrieval in RAG systems as a context assembly problem instead of an easy ranking one. A structure-aware and diversity-aware context bubble is proposed for building compact, auditable context packs under strict token budgets. It is found from results that document structure and diversity are first-class signals for efficient context construction. Explicitly modelling these signals makes it possible to achieve greater coverage, reduced redundancy and enhanced correctness of answers than flat top-K retrieval strategies. Finally, it is demonstrated that an auditable, transparent context-building approach outperforms opaque selection mechanisms in enterprise settings, where trust, control, and cost efficiency are crucial. Context Bubbles give a viable base for constructing dependable, comprehensible retrieval pipelines of actual-world RAG applications.

Although the proposed framework of context bubble has excellent empirical performance and interpretability, there are still some limitations that underlie future research. Chunk identifiers are created in a sequential manner and not in a deterministic way, restricting long-term reproducibility, incremental indexing, and integrating reuse as document collections change. Candidate generation is largely lexical, and is precise and auditable, but is sensitive to paraphrasing and semantic variation across heterogeneous document formats like narrative PDFs and free-form text. Lexical overlap thresholds are used to control diversity, and these values approximate redundancy but inadequately reflect more profound semantic overlap. Provenance is explicit on both sheet level and row level on Excel and page level on PDFs, but more expressive, modality-neutral citation anchors and massive labelled assessment across multiple tasks are open problems. Sealing these gaps, in terms of tokenizer-sensitive budgeting, deterministic chunk identities, multimodal chunking policies, hybrid lexical-semantic retrieval and better provenance model, will be the subject of our forthcoming work.

## Appendix A: Retrieval trace schema

### A1: Retrieval trace fields

| Field | Type | Description |
|---|---|---|
| chunk_id | string | Unique chunk identifier |
| sheet_name | string | Structural section / Excel sheet |
| bucket | string | Section bucket used for budget $\rho_s$ |
| row_number | int | Row index within sheet |
| token_count | int | Estimated tokens $t_i$ |
| tf | float | Lexical term frequency $tf(c_i,q)$ |
| boost | float | Structural prior ($\pi(s_i)$) + keyword boosts) |
| len_penalty | float | Length penalty $1 / (1 + t_i / \tau)$ |
| score_raw | float | tf + boost |
| score_final | float | Final score after penalty |
| stage | string | Pipeline stage (scoring, budget, overlap, etc.) |
| step_decision | string | pass / fail at stage |
| final_decision | string | selected / rejected |
| final_reason | string | budget_exceeded, too_redundant, passed_all_gates |
| step_details | JSON | Stage-specific decision details |

### A2: Example trace row abridged

| chunk_id | stage | overlap | threshold δ | final_decision |
|---|---|---|---|---|
| X000918 | overlap_gate | 0.20 | 0.55 | selected |

This trace will make it possible to deterministically and auditably rebuild each selection decision.

## Appendix B: Multi query evaluation set

### B1: Query categories

| Category | Description |
|---|---|
| Narrow fact | Single-field lookup (e.g., warranty, client name) |
| Multi-facet | Requires multiple sections (scope + materials) |
| Cross-sheet | Requires evidence across sheets |
| Table-dependent | Requires reading structured rows |

**B2: Query set composition**

| Category | Queries |
|---|---|
| Narrow fact | 6 |
| Multi-facet | 6 |
| Cross-sheet | 6 |
| Table-dependent | 7 |
| **Total** | **25** |

**Table B3: Example queries**

| Category | Example |
|---|---|
| Narrow fact | "warranty period" |
| Multi-facet | "scope of work" |
| Cross-sheet | "below grade scope and products" |
| Table-dependent | "pricing line items for waterproofing" |

The same pipeline of token budget, redundancy threshold and section budgets is used for all the queries.

**Data availability**

Data will be provided on a reasonable request

**Funding**

This work has been funded by Eye Dream Pty Ltd, 302/50 Murray St., Pyrmont, Sydney, NSW 2000, Australia and Bravada Group, Melbourne,3134, Melbourne, Australia